\documentclass[sigconf,screen,nonacm]{acmart}

\usepackage{booktabs}
\usepackage{epsfig}
\usepackage{subfigure}
\usepackage{enumitem}
\usepackage{multirow}
\usepackage{algorithm}
\usepackage{algorithmicx}
\usepackage{algpseudocode}
\usepackage{ifpdf}
\ifpdf
  
\else
\fi

\newcommand{\eg}{\textit{e}.\textit{g}.}
\newcommand{\ie}{\textit{i}.\textit{e}.}

\usepackage{color}
\definecolor{cyan}{cmyk}{1,0,0,0}
\definecolor{darkgreen}{rgb}{0,0.5,0}
\definecolor{orange}{rgb}{1,0.5,0}
\definecolor{magenta}{cmyk}{0,1,0,0}
\definecolor{darkyellow}{cmyk}{0,0,0.75,0}
\definecolor{gray}{rgb}{0.8,0.8,0.8}
\usepackage{hyperref}
\hypersetup{colorlinks=true, linkcolor=cyan, filecolor=blue, urlcolor=red, citecolor=green,}

\graphicspath{{figures/},{pairs/}}

\usepackage{fancyhdr}
\renewcommand\footnotetextcopyrightpermission[1]{}
\AtBeginDocument{%
  \providecommand\BibTeX{{%
    \normalfont B\kern-0.5em{\scshape i\kern-0.25em b}\kern-0.8em\TeX}}}

\copyrightyear{2022} 
\acmYear{2022} 
\setcopyright{acmlicensed}
\acmConference[MM '22]{Proceedings of the 30th ACM International Conference on Multimedia}{October 10--14, 2022}{Lisbon, Portugal}
\acmBooktitle{Proceedings of the 30th ACM International Conference on Multimedia (MM '22), October 10--14, 2022, Lisbon, Portugal}
\acmPrice{15.00}
\acmDOI{10.1145/3503161.3548186}
\acmISBN{978-1-4503-9203-7/22/10}

\acmSubmissionID{mmfp1806}

\begin{document}
\setcitestyle{numbers,sort&compress}

\title[]{Learnability Enhancement for Low-light Raw Denoising: \\ Where Paired Real Data Meets Noise Modeling}

	\author{Hansen Feng}
	\affiliation{%
		\institution{Beijing Institute of Technology}
		\city{Beijing}
		\country{China}}
	\email{fenghansen@bit.edu.cn}
	
	\author{Lizhi Wang}
	\authornote{Corresponding author: Lizhi Wang}
	\affiliation{%
		\institution{Beijing Institute of Technology}
		\city{Beijing}
		\country{China}}
	\email{wanglizhi@bit.edu.cn}
	
	\author{Yuzhi Wang}
	\affiliation{%
		\institution{Megvii Technology}
		\city{Beijing}
		\country{China}}
	\email{wangyuzhi@megvii.com}
	
	\author{Hua Huang}
	\affiliation{%
		\institution{Beijing Normal University}
		\city{Beijing}
		\country{China}}
	\email{huahuang@bnu.edu.cn}

\renewcommand{\shortauthors}{}

\renewcommand{\thefootnote}{}

\begin{abstract}
    Low-light raw denoising is an important and valuable task in computational photography where learning-based methods trained with paired real data are mainstream.
    However, the limited data volume and complicated noise distribution have constituted a learnability bottleneck for paired real data, which limits the denoising performance of learning-based methods.
    To address this issue, we present a learnability enhancement strategy to reform paired real data according to noise modeling.
    Our strategy consists of two efficient techniques: shot noise augmentation (SNA) and dark shading correction (DSC).
    Through noise model decoupling, SNA improves the precision of data mapping by increasing the data volume and DSC reduces the complexity of data mapping by reducing the noise complexity.
    Extensive results on the public datasets and real imaging scenarios collectively demonstrate the state-of-the-art performance of our method. Our code will be released at: \url{https://github.com/megvii-research/PMN}.

\end{abstract}




\maketitle

\section{Introduction}\label{sec:Introduction}
Low-light denoising is a fundamental problem for the increasingly widespread computational photography, especially on mobile devices~\cite{ECCV20/Yuzhi}. Learning-based methods have made great progress in recent years and became the mainstream solution for the problem~\cite{TPAMI21/LowLightSurvey}.
The standard paradigm is to learn a mapping between paired real data, \ie, the low-light noisy image and its clean counterpart.

Unfortunately, the mapping between paired real data is difficult to be learned and thus suffers from fragile learnability, which limits the denoising performance.
On one hand, the real noise follows a complicated distribution considering the imaging process of image sensors, leading to the high complexity of the data mapping. On the other hand, the data volume is limited by the physical environment, resulting in the low precision of the data mapping.
The complicated noise distribution and limited data volume constitute a mapping dilemma of paired real data in low-light denoising.

Recent works try to solve the mapping dilemma by synthesizing data according to noise modeling instead of employing the paired real data~\cite{CVPR19/Unprocess,TPAMI21/ELD}. Such methods can synthesize new noisy and clean image pairs for learning the data mapping. However, since some parts of the noise (\ie, read noise) are far from accurately modeled, the synthetic data would inescapably deviate from the real data, which loses effectiveness in real imaging scenarios. Thus, the fragile learnability is still a bottleneck of paired real data in low-light denoising.

In this paper, we propose a learnability enhancement strategy for low-light raw denoising by reforming paired real data according to noise modeling.

Our first observation is that the shot noise is only related to the clean image and can be accurately modeled with the Poisson distribution~\cite{SPIE85/CCD,TPAMI94/CCD}.
We propose the Shot Noise Augmentation (SNA) method to increase the data volume of paired real data.
SNA first combines the real data and the shot noise model to synthesize new noisy-clean data pairs and then puts them into the training process to improve the mapping precision.
Benefiting from the increased data volume, the mapping can promote the denoised images with clear texture.

Our second observation is that the dark shading keeps temporal stable and can be modeled with a pixel-wise bias~\cite{TIP14/FPNR,darkshading}.
We propose the Dark Shading Correction (DSC) method to reduce the complexity of real noise distribution. 
DSC first calibrates the dark shading and then corrects it in the noisy image to reduce the mapping complexity.
Benefiting from the reduced noise complexity, the mapping can promote the denoised images with exact colors.


The ideas of SNA and DSC originate from the integration of paired real data and noise modeling which are assumed as two parallel directions for low-light raw denoising.
Through noise model decoupling, our method achieves higher learnability.
Extensive experiments on the public SID dataset~\cite{CVPR18/SID}, ELD dataset~\cite{TPAMI21/ELD}, as well as real imaging scenarios, demonstrate the state-of-the-art performance of our method.

Our main contributions are summarized as follows:

\begin{itemize}[leftmargin=*]
    \item We light the idea of learnability enhancement for low-light raw denoising by utilizing noise modeling to reform paired real data.
    \item We increase the data volume of paired real data with a novel shot noise augmentation method, which improves the precision of data mapping through decoupling the real noise into shot noise and read noise.
    \item We reduce the complexity of real noise distribution with a novel dark shading correction method, which reduces the complexity of data mapping through decoupling the read noise into temporal stable noise and temporal variant noise.
    \item We demonstrate the superior performance of our method on the public datasets and real imaging scenarios.
\end{itemize}

\section{Related Works}\label{sec:RelatedWorks}
Low-light denoising is important in the computational photography, which is widely needed in mobile photography~\cite{TOG16/HDR+,TOG19/HDR+3,ECCV20/Yuzhi}, astronomy~\cite{astronomy}, remote sensing~\cite{remote} and microscopic imaging~\cite{microscopic}.
Classical denoising methods usually rely on image priors such as smoothness~\cite{wavelet,TV}, self-similarity~\cite{NLM,BM3D,VBM4D}, sparsity~\cite{elad2006image, K-SVD}, and low rank~\cite{WNNM}.
Instead of pre-setting image prior, learning-based methods directly learn the mapping from the noisy image to its clean counterpart (paired real data) via deep neural networks.
Recent works demonstrate that learning-based methods have been far superior to classical methods in denoising performance~\cite{CVPR18/SID, ICCV19/DRV, RViDeNet}.
Although simple and powerful, these learning-based methods trained with paired real data are often suffering from fragile learnability.
The complicated data mapping and limited data volume have restricted the development of denoising performance.

A line of research has focused on improving the realism of noise modeling to circumvent the learnability bottleneck of paired real data.
The sensor noise is usually divided into ``shot noise" and ``read noise".
By modeling shot noise with Poisson and read noise with Gaussian, Poisson-Gaussian model is widely employed to characterize the real noise in raw denoising~\cite{P-G}.
In order to match the noise model to the practical application, various calibration~\cite{EMVA1288,TPAMI94/CCD,P-G} and variance-stabilizing transformations~\cite{TIP11/VST,TIP13/VST} have been proposed to ensure the noise model can work on different cameras under any conditions.
Based on the refinement and summary of noise modeling, \citeauthor{CVPR20/ELD} proposes ELD, a physics-based noise modeling for extreme low-light photography, whose results are on par with the results trained with paired real data.
Unfortunately, ELD still exists limitations since its assumptions are only based on temporal noise (i.i.d.), deviating from real spatial noise (\eg, fixed pattern noise).
As a result, the visual results of ELD are still suffering from residual pattern noise, which indicates that real data is still essential.
Thus, noise modeling can not address the learnability bottleneck of paired real data itself.

Nonetheless, noise modeling has shown great potential.
Since the physical properties of the photoelectric reaction is well-defined, the analysis of shot noise is reliable~\cite{SPIE85/CCD,TPAMI94/CCD,EMVA1288}. In contrast, the read noise is so complicated that there is no consensus in the imaging community yet~\cite{SPIE04/imaging,TED07/CMOS,EMVA1288,arxiv2014/CMOS,TIP14/FPNR}.
In other words, it is impossible to exactly extract and model all kinds of noise sources in the camera.
Due to this reason, SFRN adopts reliable shot noise for noise modeling and samples complicated read noise from real dark frames, which has liberated the amount of paired real data~\cite{ICCV21/SFRN}.
Although SFRN combines the advantages of noise modeling and real data, however, it also inherits the complexity of real data, making learning data mapping a challenging process.

We find that the high complexity of the data mapping is mainly contributed by dark shading~\cite{darkshading}.
Unfortunately, SFRN is incompatible with dark shading due to its implementation, so it is necessary to find another approach to combine real data and noise modeling.
In order to address both complicated data mapping and limited data volumes, we utilize noise modeling to reform paired real data through noise model decoupling. Solving fragile learnability leads to significant improvements in denoising performance on real data, particularly under the condition of low light.

\section{Method}\label{Methods}
    In this section, we firstly introduce the overview of our framework, which includes the principle and procedure of our learnability enhancement strategy. Second, we introduce our Shot Noise Augmentation based on real noise decoupling, which can improve the precision of data mapping. Finally, we introduce our Dark Shading Correction based on read noise decoupling, which can reduce the complexity of data mapping.

\subsection{Framework}
\label{Framework}
The mapping between paired real data is difficult to be learned, which limits the denoising performance of learning-based methods. 
The limited data volume and complicated noise distribution are the main culprits that lead to the fragile learnability of paired real data.
These two points are difficult to be solved since changes in the real paired data usually break the real noise distribution, thereby making the mapping inaccurate.
Our learnability enhancement strategy reforms paired real data through noise model decoupling, which can enhance the learnability of the mapping without breaking real noise distribution.

\begin{figure}[t!]
    \begin{center}
      \includegraphics[width=\columnwidth]{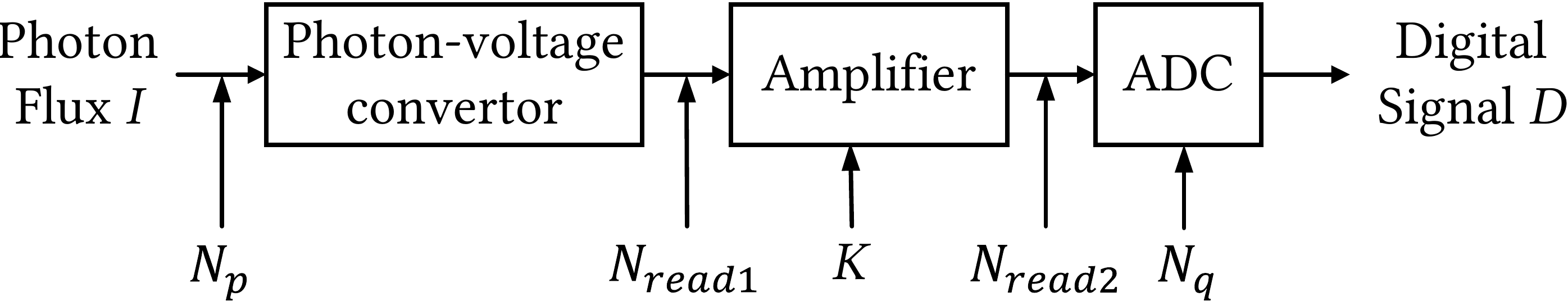}
    \end{center}
    \caption{Simplified imaging model of the sensor.}
    \label{fig:Formation}
\end{figure}
\begin{figure*}[t!]
    \begin{center}
      \includegraphics[width=\textwidth]{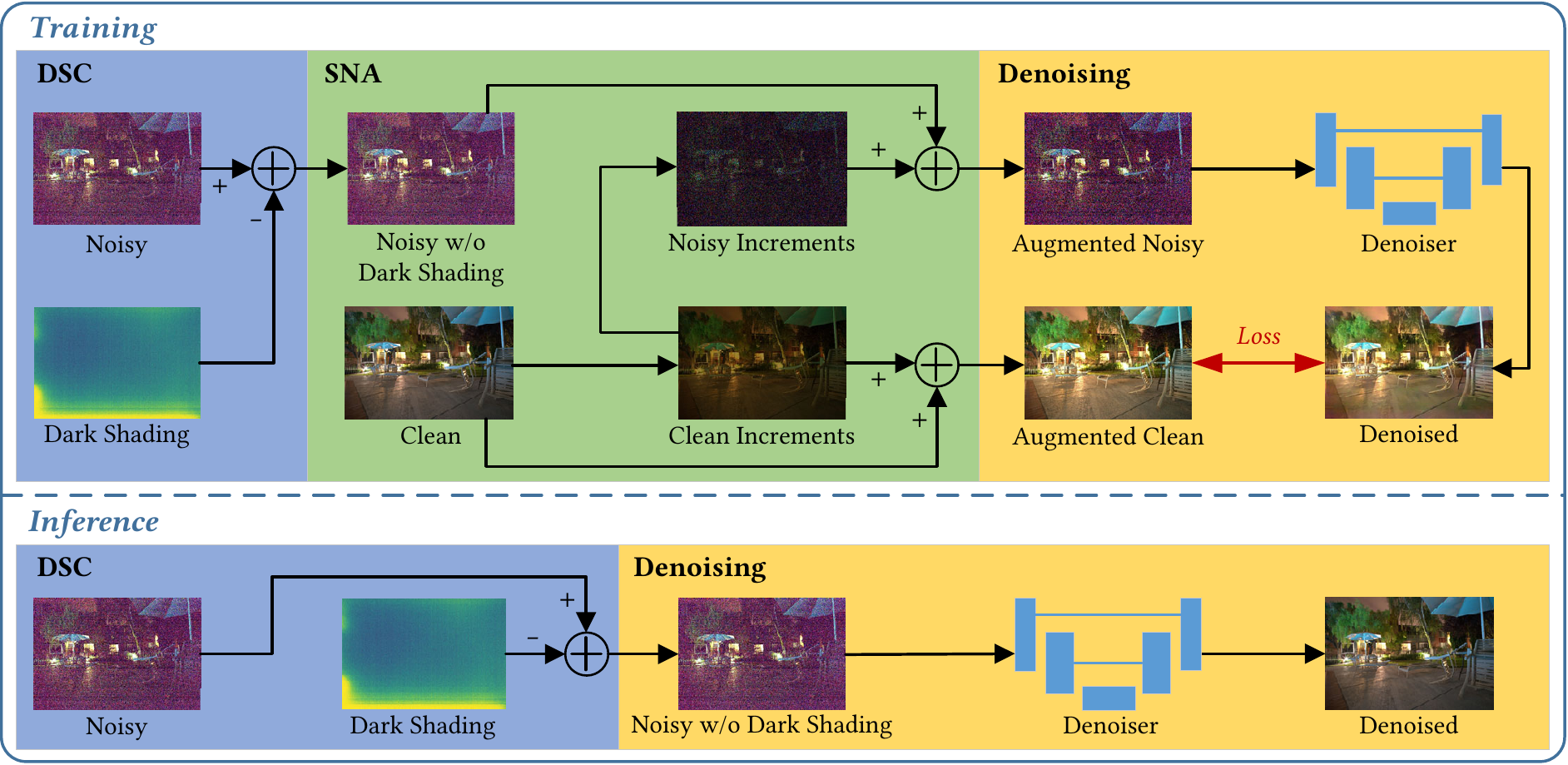}
    \end{center}
    \caption{Overview of our framework.}
    \label{fig:framework}
\end{figure*}

To clearly explain our work, we will briefly introduce the imaging model of the sensor.
As shown in Fig.~\ref{fig:Formation}, in the camera electronics, the voltage signal is converted from the accumulated photon-electron charge units, before it is amplified and finally converted into digital signal $D$ by the analog-digital convertor~(ADC)~\cite{EMVA1288}.
For a raw image captured by a sensor, we model the noise formation process as
\begin{equation}\label{eq:raw}
    D = K(I + N_p + N_{read1}) + N_{read2} + N_q
\end{equation}
where $K$ represents overall system gain, $I$ is the number of photo-electrons excited by the scene radiance, $N_p$ is the signal-dependent photon shot noise, $N_{read1}$ and $N_{read2}$ are read noise before and after amplifier respectively, and $N_q$ is the quantization error between the analog input voltage to the ADC and the output digitized value.

Among the image noise components, $K$ and $N_p$ can be estimated by the Photon Transfer method~\cite{SPIE85/CCD, EMVA1288}. However, for the rest noise components, the physical complexity leads to no consensus in the community as we have described in Section~\ref{sec:RelatedWorks}. Without loss of generality, we use $N_{read}$ to refer to other signal-independent noise dominated by read noise. 

The real data is a complex whole, so it is difficult to decompose it into different components according to the noise model. However, the fitting target of the learning-based methods is the reliable data mapping among the paired real data, rather than the real paired data itself. Thus, we can augment and decouple the real data as long as these operations do not break the real noise distribution.
Therefore, we decouple the real noise model as
\begin{equation}\label{eq:raw-simple}
    D = K (I + N_{p}) + N_{read}
\end{equation}
where $N_{read} = KN_{read1} + N_{read2} + N_q$.

Based on this decoupling, we can enhance the learnability according to the noise characteristics of shot noise and read noise respectively.

Our strategy consists of two parts.
SNA is a data augmentation method for shot noise. We continuously synthesize new noisy-clean data pairs by augmenting the shot noise to increase the limited data volume.
DSC is the decoupling for the read noise. We first calibrate the temporal stable component (dark shading) and then correct it in paired real data to reduce the noise complexity.
Both SNA and DSC do not change the noise model, so our strategy can enhance the learnability of the mapping without breaking the real noise distribution.

Our method is easier to understand from a mapping perspective.
Now we suppose that there is a noisy image $D_{n}$ and a corresponding clean image $D_{c}$. 
The relationship between them satisfies
\begin{equation}\label{eq:Dnoisy}
    D_{n} \sim \mathcal F(D_{c})
\end{equation}
where $\mathcal F$ donates the noise model, $D_{n}$ follows Eq.~\eqref{eq:raw-simple}, $D_{c}$ can be modeled as $KI$.

In mathematics, the denoising task can be abstracted as its inverse function $\mathcal F^{-1}$, given by
\begin{equation}\label{eq:NRMapping}
    \mathcal F^{-1}:  D_{n} \rightarrow D_{c}
\end{equation}

Combing Eq.~\eqref{eq:raw-simple} and Eq.~\eqref{eq:NRMapping}, we obtain the denoising mapping represented by the denoising model $\mathcal F^{-1}$ as
\begin{equation}\label{eq:NRmapping_old}
    \mathcal F^{-1}:  K(I + N_p) + N_{read} \rightarrow KI
\end{equation}

By applying our learnability enhancement, the enhanced mapping relationship in our framework can be represented as
\begin{equation}\label{eq:ours}
        \mathcal F_{LE}^{-1}: \underbrace{K (I + N_{p}) + \Delta N}_{SNA} + \underbrace{(N_{read} - D_{ds})}_{DSC} \rightarrow \underbrace{KI + \Delta D}_{SNA}
\end{equation}
where $\mathcal F_{LE}^{-1}$ is the mapping applying our learnability enhancement, $D_{ds}$ is the dark shading produced by DSC, $\Delta N$ and $\Delta D$ are noisy and clean signal increments produced by SNA respectively. The details of these variables will be covered in subsequent subsections.

The framework of our learnability enhancement is shown in Fig.~\ref{fig:framework}. For training, we first correct the dark shading hiding in the noisy raw image by DSC. Then we augment the clean image and the noisy image to obtain new data pairs by SNA. Finally, We use augmented noisy image and augmented clean image to train a denoiser with UNet-like~\cite{Unet} architecture. For inference, we just denoise the noisy image after correcting dark shading with the trained denoising model.





 \subsection{Shot Noise Augmentation}
  \label{SNA}
    \paragraph{\textbf{The principle of SNA}}
    SNA is a data augmentation method based on photon shot noise modeling.
    Due to the quantum nature of light and the uncertainty of the collected photon numbers, $(I + N_p)$ for all pixels follow the Poisson distribution
    \begin{equation}\label{eq:possion}
        (I + N_p) \sim{\mathcal P(I)}
    \end{equation}
    where $\mathcal P$ denotes the Poisson distribution.
    Variables following Poisson distributions satisfies the additive property. If ${X_{1}\sim \mathcal{P} (\lambda _{1})}$ and ${X_{2}\sim \mathcal{P} (\lambda _{2})}$ are independent, then ${X_{1} + X_{2}\sim \mathcal{P} (\lambda_{1} + \lambda_{2})}$.

    Now we suppose that there is a clean signal increments $\Delta D$ and our target is to find a corresponding noisy signal increments $\Delta N$ satisfying
    \begin{equation}\label{eq:Delta}
        (D_{n} + \Delta N) \sim \mathcal F(D_{c} + \Delta D)
    \end{equation}

    For paired real data, the clean image $D_{c}$ is known.
    According to Eq.~\eqref{eq:raw-simple} and Eq.~\eqref{eq:possion}, the noise model $\mathcal F(D_{c})$ is given by
    \begin{equation}\label{eq:noise model}
        \mathcal F(D_{c}) = K \mathcal P(\frac{D_{c}}{K}) + \mathcal F_{read}
    \end{equation}
    where $\mathcal F_{read}$ is the distribution of $N_{read}$, which is unknown.

    According to the additivity of Poisson distribution, there is
    \begin{equation}\label{eq:SNA}
        \begin{aligned}
            \mathcal F(D_{c} + \Delta D) &= K \mathcal P(\frac{D_{c} + \Delta D}{K}) + \mathcal F_{read} \\
            &= K \mathcal P(\frac{D_{c}}{K}) + K \mathcal P(\frac{\Delta D}{K}) + \mathcal F_{read} \\
            &= \mathcal F(D_{c}) + K \mathcal P(\frac{\Delta D}{K})\\
        \end{aligned}
    \end{equation}

    Bring Eq.~\eqref{eq:Dnoisy} and Eq.~\eqref{eq:SNA} into Eq.~\eqref{eq:Delta}, then we can obtain
    \begin{equation}
        \label{eq:deltaN}
        \Delta N \sim K \mathcal P(\frac{\Delta D}{K})
    \end{equation}

    At last, we add clean signal increments $\Delta D$ to the clean image and noisy signal increments $\Delta N$ to the noisy image, so that we can achieve data augmentation without breaking the real noise distribution. Since noisy signal increments, $\Delta N$ only need to follow Eq.~\eqref{eq:deltaN}, even if clean signal increments $\Delta D$ are consistent, we can obtain new noisy data every time we randomly sample.

    By increasing the data volume through SNA, the model can precisely fit the mapping relationship among the paired real data, which promotes the denoised images with clear texture.

    \paragraph{\textbf{The procedure of SNA}}
    The augmentation procedure has been shown in Procedure~\ref{alg:SNA}.
    \floatname{algorithm}{Procedure}
    \begin{algorithm}[t!]{}
    	\caption{Shot Noise Augmentation}
    	\label{alg:SNA}
    	\begin{algorithmic}
    		\Require $D_{c}, D_{n}$
    		\Ensure $D_{c}^*, D_{n}^*$
    		\Function{parameter sampling}{$\mu, \sigma$}
    		\State $\epsilon_{g} \sim \mathcal N(\mu+1, \sigma),\quad\epsilon_{r}, \epsilon_{b} \sim \mathcal N(1, \sigma)$
    		\State $Gain_{g} \gets clip(\epsilon_{g})_1^{4\mu+1}$
    		\State $Gain_{r} \gets clip(Gain_{g} \cdot \epsilon_{r})_1^{4\mu+1}$
    		\State $Gain_{b} \gets clip(Gain_{g} \cdot \epsilon_{b})_1^{4\mu+1}$
    		\State $Gain \gets (Gain_{r}, Gain_{g}, Gain_{b})$
    		\State \Return{$Gain$}
    		\EndFunction
    		\State $Gain \gets$ \Call{parameter sampling}{$\mu, \sigma$}
    		\State $D_{c}^* \gets$ $D_{c} \cdot Gain$
    		\State $\Delta D \gets D_{c}^* - D_{c}$
    		\State $\Delta N \sim K \mathcal P(\frac{\Delta D}{K})$
    		\State $D_{n}^* \gets D_{n} + \Delta N$
    	\end{algorithmic}
    \end{algorithm}
    
    Firstly, we randomly sample a set of parameters to gain the clean image $D_{c}$ in order to obtain an augmented clean image $D_{c}^*$. Secondly, we obtain the clean signal increments $\Delta D$ by computing the difference of clean image $D_{c}$ and augmented clean image $D_{c}^*$. Then, we synthesize noisy signal increments $\Delta N$ according to the clean signal increments $\Delta D$. Finally, we add the synthesized noisy signal increments $\Delta N$ to the real noisy image $D_{n}$ to obtain an augmented noisy image $D_{n}^*$. The augmented noisy image $D_{n}^*$ and augmented clean image $D_{c}^*$ will constitute a new noisy-clean image pair.

    It is free to design the specific parameter sampling strategy for the synthesis of clean signal increments $\Delta D$, however, some basic principles still need to be followed:
    (1) clean signal increments $\Delta D$ should be non-negative to ensure that Poisson sampling can be applied;
    (2) it is necessary to ensure that SNA will not introduce obvious color bias and large-scale over-exposure.
    
    Based on the above principles, we choose to synthesize clean signal increments $\Delta D$ based on clean image $D_{c}$. We only augment the ratio of color channels and global brightness in order to simulate the real scene, which refers to the white balance modeling~\cite{CVPR19/Unprocess}.
    First, we randomly initialize factors $\epsilon_{r},\epsilon_{g},\epsilon_{b}$ that follow a Gaussian distribution, where $\mu$ and $\sigma$ are the parameters of the Gaussian distribution.
    Then, we randomly sample the gain of the green channel $Gain_{g}$ based on the clipped $\epsilon_{g}$.
    Next, we select the gain of the red channel $Gain_{r}$ and blue channel $Gain_{b}$ based on the random factors $\epsilon_{r}, \epsilon_{b}$ and the gain of the green channel $Gain_{g}$.
    Finally, the gain parameters $(Gain_{r}, Gain_{g}, Gain_{b})$ is used to gain different color channels, and they are all clipped to the appropriate value range to meet the basic design principles.


    It is important to note that the data augmentation for raw denoising is special.
    In addition to the difficulty caused by the Bayer pattern of the raw image itself~\cite{CVPRW19/fhq}, the changes in signal amplitude usually break the real noise distribution in the data, which can produce a domain gap between training and application.
    Thus, raw denoising methods can only use limited data augmentation methods (\eg, rotation, flipping) cautiously, which is useless for the lack of data volume. Moreover, these data augmentation methods cannot increase the diversity of noise signals, which is of limited help for fitting the data mapping.
    Thus, the SNA, which can increase noise diversity without breaking the noise distribution, is both novel and important in raw denoising.
    
  \subsection{Dark Shading Correction}
  \label{DSC}
    \paragraph{\textbf{The principle of DSC}}
    In this part, we will introduce the definition and mechanism of dark shading and clarify the importance of correcting the dark shading from the noise model for low-light image denoising.

    All properties of an array of pixels will vary from pixel to pixel in the sensor~\cite{EMVA1288}.
    According to the definition in~\cite{darkshading}, we follow the mainstream term ``dark shading" in the industry to describe such spatial inhomogeneity from a practical perspective.
    In general, dark shading represents the temporal stable component in the read noise, which can be regarded as the union of black level error (BLE)~\cite{ICME21/RethinkNM, TPAMI21/ELD, nakamura2017image} and fixed pattern noise (FPN)~\cite{EMVA1288,TIP01/FPNR,TIP14/FPNR,arxiv2014/CMOS,2011/CMOS}.
    BLE is the spatial invariant bias of dark shading while FPN is considered a spatial variant pattern.

    Dark shading is closely related to dark current, which means it also suffers from temperature dependency~\cite{2011/CMOS,EMVA1288}. Fortunately, thanks to the development of circuit~\cite{AICSP19/CDS,PRIME14/CDS}, mainstream modern sensors no longer suffer the impacts of temperature dependency within the normal operating temperature range. Thus, dark shading can be treated as temporal stable under the same camera settings and in similar environments.
    
    Most denoising works follow the ideal assumption that sensor noise is considered to be temporal stable noise with zero means.
    In the ideal assumption, the expectation of the noisy image $\mathbb E(D_{n})$ is same as the mapping target $D_{c}$ in mathematics.
    However, in the real noise, the expectation of read noise $\mathbb E(N_{read})$ is the non-zero dark shading, which breaks the ideal assumption.
    If spatial variant dark shading is considered in real noise, the mapping between $\mathbb E(D_{n})$ and clean image $D_{c}$ will become a patch-wise non-injective mapping.
    Non-injective mapping has no inverse mapping, so it is virtually impossible to fit an accurate mapping for denoising through patch-wise training due to dark shading. Thus, the mapping dilemma between $\mathbb E(D_{n})$ and clean image $D_{c}$ will lead to complicated data mapping from noisy image $D_{n}$ to clean image $D_{c}$.

    \begin{figure}[t!]
        \includegraphics[width=\columnwidth]{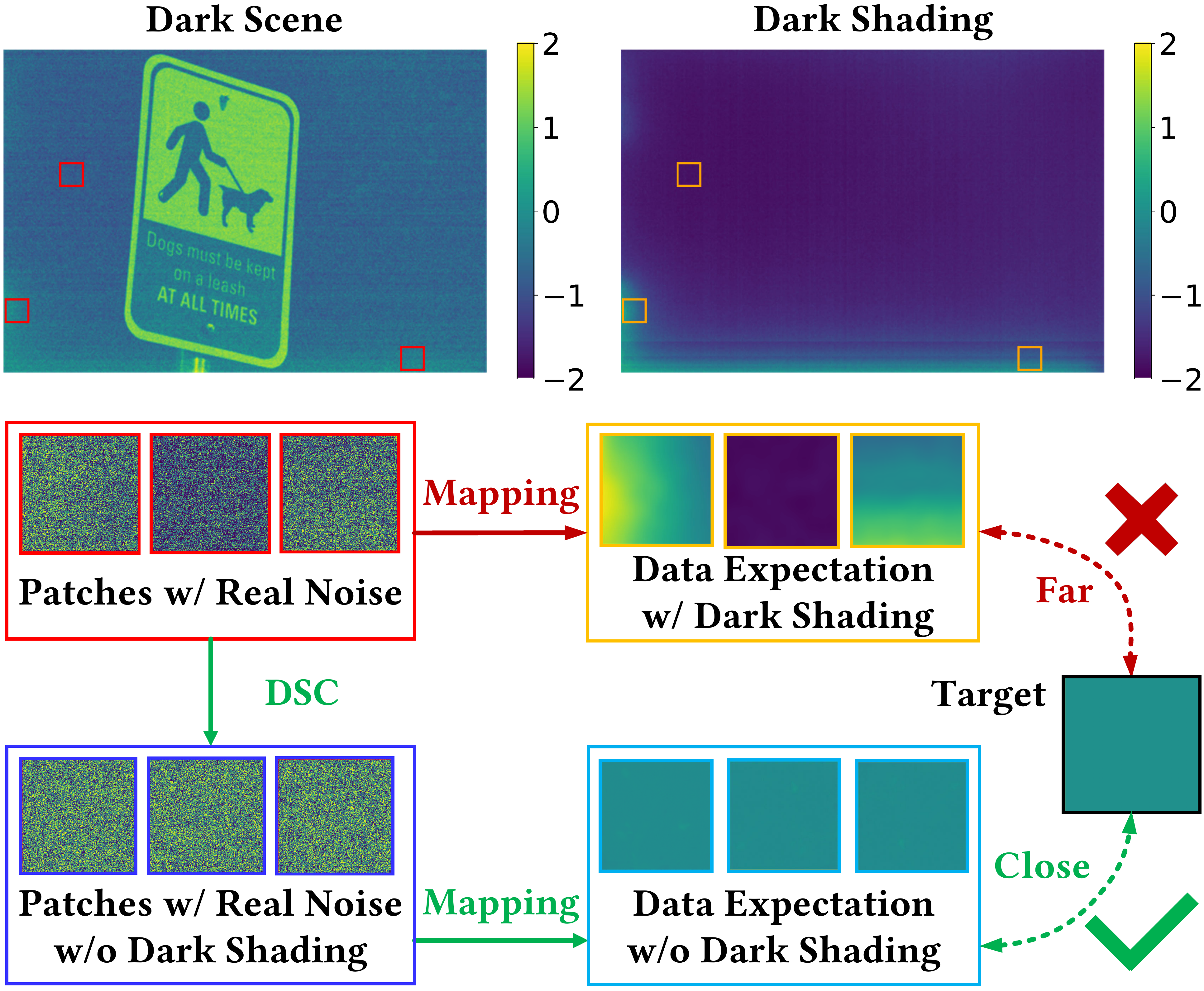}
        \caption{An example of the mapping dilemma considering dark shading. We have converted the color space and limited the range of values for viewing.}
        \label{fig:Mapping}
    \end{figure}
    
    \begin{figure}[t!]
        \includegraphics[width=\columnwidth]{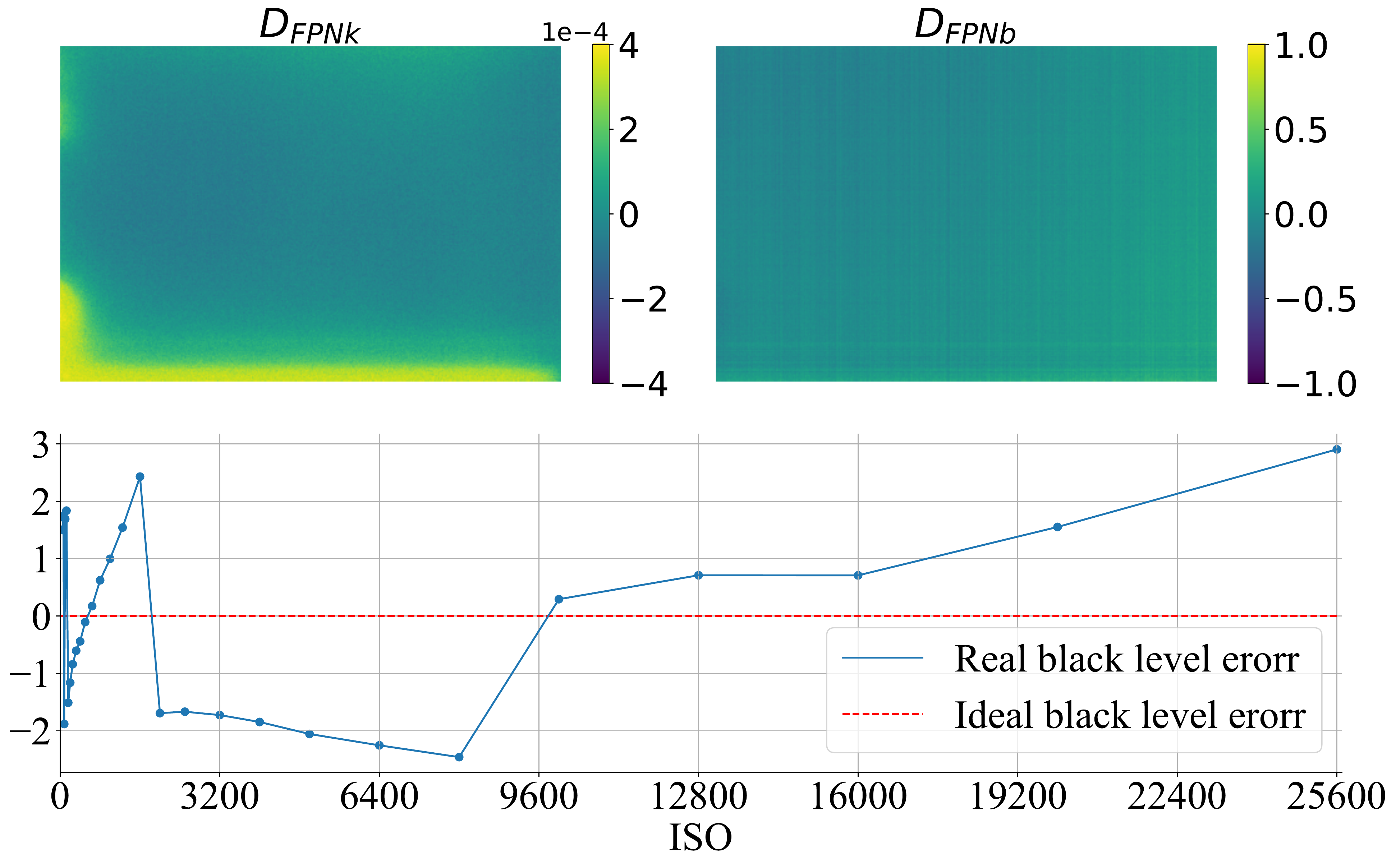}
        \caption{The calibration results of dark shading.}
        \label{fig:FastDSC}
    \end{figure}

    We will intuitively show this mapping dilemma with the help of an example shown in Fig.~\ref{fig:Mapping}.
    The dark scene represents an image collected in a dark environment and the corresponding dark shading is shown on the right.
    The red patches are cropped from the dark scene and the yellow patches are corresponding dark shading at the same location. The blue patches are the result of applying DSC to the red patches.
    The sky blue patches represent the data expectation of the noisy image without dark shading.
    The black target patch corresponds to the clean counterpart of the dark scene patches, which are all zeros in this example.

    For the data containing zero-mean noise in the ideal assumption, the mapping target is the expectation of the data. However, when such a mapping is applied to the real noise, the mapping target contains the corresponding dark shading.
    If we directly force the mapping target to be a clean target (black patch in this example), it will be equivalent to taking the dark red path, where the mapping dilemma will arise. Since the receptive field of the convolutional neural network is limited, without global position information, we cannot treat dark shading at different positions differently, nor can we use the same mapping to correct all the spatial patterns caused by dark shading.
    Conversely, if we apply DSC to the real noise first, it will be equivalent to taking the light green path, where the data mapping will no longer be disturbed by the dark shading. Since the above mapping dilemma no longer exists, the complexity of data mapping will be greatly reduced.

    Previous learning-based methods believe that convolutional neural networks, a collection of local operations, can be used to fit data pairs to remove pattern noise caused by dark shading, but this is practically impossible in principle.
    Thus, it is extremely important to correct the spatial variant dark shading before denoising in a learning-based denoising method.
    
    By reducing the mapping complexity through DSC, the mapping relationship among the paired real data will be easier to learn, which promotes the denoised images with exact colors.

\begin{table*}[t!]
    \caption{Quantitative results (PSNR/SSIM) of different methods on the ELD dataset and SID dataset with different exposure ratios. The {\color{red}red} color indicates the best results and the {\color{blue}blue} color indicates the second-best results.}
    \label{tab:compare}
    {%
        \begin{tabular}{ccccccc}
        \toprule
        {\makebox[0.075\textwidth][c]{}} & {\makebox[0.075\textwidth][c]{}} & 
        {\makebox[0.11\textwidth][c]{P-G}} & 
        {\makebox[0.11\textwidth][c]{ELD}} & 
        {\makebox[0.11\textwidth][c]{SFRN}} & 
        {\makebox[0.11\textwidth][c]{Paired data}} & 
        {\makebox[0.11\textwidth][c]{Ours}} \\ \cline{3-7} 
        \multirow{-2}{*}{Dataset} & \multirow{-2}{*}{Ratio} & PSNR / SSIM & PSNR / SSIM & PSNR / SSIM & PSNR / SSIM & PSNR / SSIM \\ \midrule
        & $\times$100 & 42.05 / 0.872 & 45.45 / 0.975 & {\color{blue} 46.02 / 0.977} & 44.47 / 0.968 & {\color{red} 46.50 / 0.985} \\
        \multirow{-2}{*}{ELD} & $\times$200 & 38.18 / 0.782 & 43.43 / 0.954 & {\color{blue} 44.10 / 0.964} & 41.97 / 0.928 & {\color{red} 44.51 / 0.973} \\ 
        \midrule
         & $\times$100 & 39.44 / 0.890 & 41.95 / 0.963 & {\color{blue} 42.29} / 0.951 & 42.06 / {\color{blue} 0.955} & {\color{red} 43.16 / 0.960} \\
         & $\times$250 & 34.32 / 0.768 & 39.44 / 0.931 & {\color{blue} 40.22 / 0.938} & 39.60 / {\color{blue} 0.938} & {\color{red} 40.92 / 0.947} \\
        \multirow{-3}{*}{SID} & $\times$300 & 30.66 / 0.657 & 36.36 / 0.911 & {\color{blue} 36.87} / 0.917 & 36.85 / {\color{blue} 0.923} & {\color{red} 37.77 / 0.934} \\ 
        \bottomrule
        \end{tabular}%
        }
\end{table*}

    \paragraph{\textbf{The procedure of DSC}}

    In this part, we will introduce the procedure of dark shading correction.
    
    Firstly, we average 400 dark frames at different ISO to obtain the calibration materials of dark shading based on the zero-mean property of the temporal noise.
    All calibration materials need to be collected in a lightless environment with an exposure time of 1/30 second.

    Since we find that FPN is clearly linear with ISO while BLE has no clear mapping relationship with ISO, we extract the dark shading by fitting the regression model after getting the calibration material
    \begin{equation}\label{eq:DS-linear}
        D_{ds} = D_{FPNk} \cdot ISO + D_{FPNb} + D_{BLE}(ISO)
    \end{equation}
    where $D_{FPNk}$ and $D_{FPNb}$ are the coefficient matrices of the FPN we need to regress in the dark shading. $D_{BLE}(ISO)$ is the BLE at a specific ISO, which can be regarded as the global average of dark shading.

    Compared with directly obtaining calibration materials at all ISO as dark shading, using a linear model to model dark shading has two advantages.
    Firstly, linear regression can be regarded as a regularization for dark shading, which can further reduce the temporal variant noise that remains in the calibrated dark shading.
    Secondly, linear regression allows us to reduce laborious data collection, which means that collecting dark frames at partial ISO can also finish calibration.
    In this work, we collect dark frames for calibration only at the ISO from the set $\{100\times2^n | n\in \mathbb{N}, 0 \le n\le 8\}$.
    
    Since the relationship between BLE and ISO is not regular, we collect a dark frame at each ISO and calculate the average of the whole image as the BLE at this ISO. The error caused by the number of frames is negligible according to our experiments. 
    

    The results of the calibration have been shown in Fig.~\ref{fig:FastDSC}. The top left image is $D_{FPNk}$ and the top right image is $D_{FPNb}$. The chart below represents the BLE at different ISO.
    In the calibration procedure, calibrated BLE is first corrected to normalize the dark shading. Then we use linear regression to fit the dark shading at different ISO to obtain $D_{FPNk}$ and $D_{FPNb}$. Finally, we can reconstruct the dark shading at any ISO by Eq.~\eqref{eq:DS-linear}.

    After obtaining the dark shading, we will correct it from the real raw images in advance.
    Same as dark frame subtraction~\cite{naiveDSC,DFS}, the purpose of DSC is to decouple the read noise model $\mathcal F_{read}$ into the temporal stable dark shading and the temporal variant noise, which will significantly reduce the complexity of the data mapping, so as to improve the learnability of the data mapping.



\section{Experiment}
    In this section, we firstly present the experimental setting including implementation details and compared methods.
    Secondly, we compare our method against prior art with quantitative and visual results on the SID dataset~\cite{CVPR18/SID} and ELD dataset~\cite{TPAMI21/ELD}.
    Then we conduct comprehensive ablation studies for an in-depth analysis of our SNA and DSC.
    Finally, we evaluate our method on a smartphone camera to demonstrate the generalization of our method.
  \subsection{Experimental Setting} \label{subsec:setting}
    \paragraph{\textbf{Implementation details}}
    We use the same UNet-like~\cite{Unet} architecture as SID and preprocess raw images in a similar way to ELD.
    We use the SID Sony set for training. The long-exposure raw bayer images are used to create synthetic data for comparison.
    The Sony set in both SID dataset and ELD dataset are used for validation.

    Our framework has been shown in Fig.~\ref{fig:framework}. DSC is applied to the noisy raw images and SNA is only used for training, there is a 1/4 probability that the training data pairs will not be augmented by SNA. The calibration is based on a SonyA7S2 camera, which has the same sensor as the public datasets but not the same camera.
    We pack the raw Bayer images into four channels, and sample 8 non-overlapped $512 \times 512$ patches with rotation and flipping augmentation each image as a batch. 
    We visualize the raw images as sRGB images for viewing through Rawpy (a Python wrapper for LibRaw) with the metadata of reference images following the existing works~\cite{CVPR18/SID, TPAMI21/ELD}.


    Our implementation is based on PyTorch.
    We train the models with 1800 epochs using Adam optimizer~\cite{Adam} and $L_1$ loss. The learning rate will vary with iterations like SGDR~\cite{ICLR17/SGDR}. The base learning rate is set to $2\times10^{-4}$ and the minimum learning rate is set to $10^{-5}$. The optimizer restarts every 600 epochs and the learning rate is halved on restarts.

    \paragraph{\textbf{Compared methods}}
    In order to demonstrate the reliability of our strategy, we compare our approach with:
    \begin{itemize}[leftmargin=*]
        \item The model trained with synthetic data based on different noise models, including Poisson-Gaussian (P-G)~\cite{P-G} and ELD~\cite{TPAMI21/ELD}, which is the classical method for low-light raw denoising.

        \item The model trained with the data combines noise modeling and sampled real noise proposed by SFRN~\cite{ICCV21/SFRN}, which is the state-of-the-art method before our work.

        \item The model trained with paired real data proposed by SID~\cite{CVPR18/SID}, for which our learnability enhancement strategy needs to serve.
    \end{itemize}

    The results of P-G, ELD, and Paired data are implemented from the code and weights released by the open-source project. Since the training code and weights of SFRN have never been released, we reproduced the results of SFRN with the help of the authors and the results of the experiments have been confirmed by the authors.
    
    \begin{figure*}[t!]
        \includegraphics[width=\textwidth]{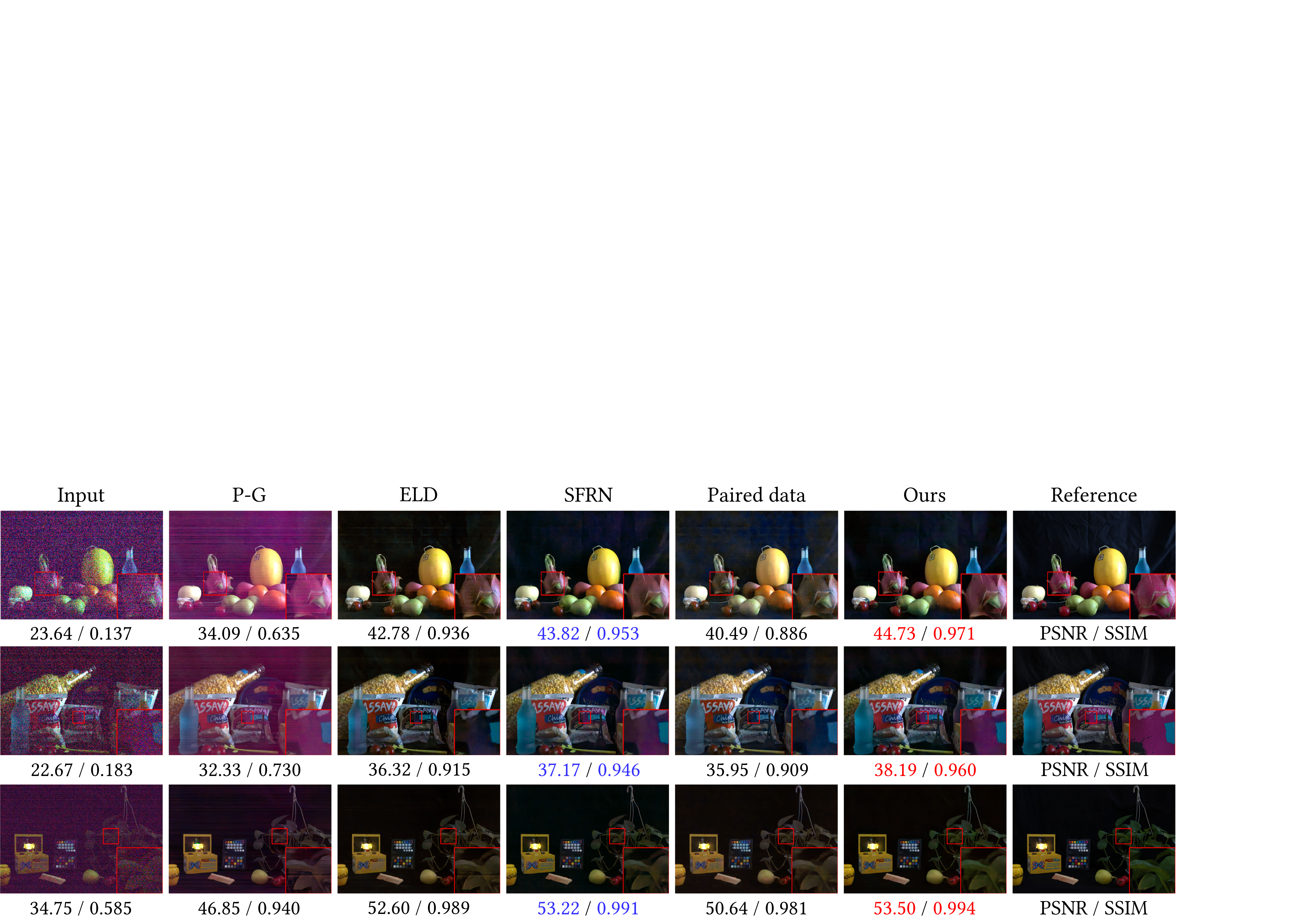}
        \caption{Raw image denoising results on images from the ELD dataset. (Best viewed with zoom)}
        \label{fig:ELD-compare}
    \end{figure*}
    \begin{figure*}[t!]
        \includegraphics[width=\textwidth]{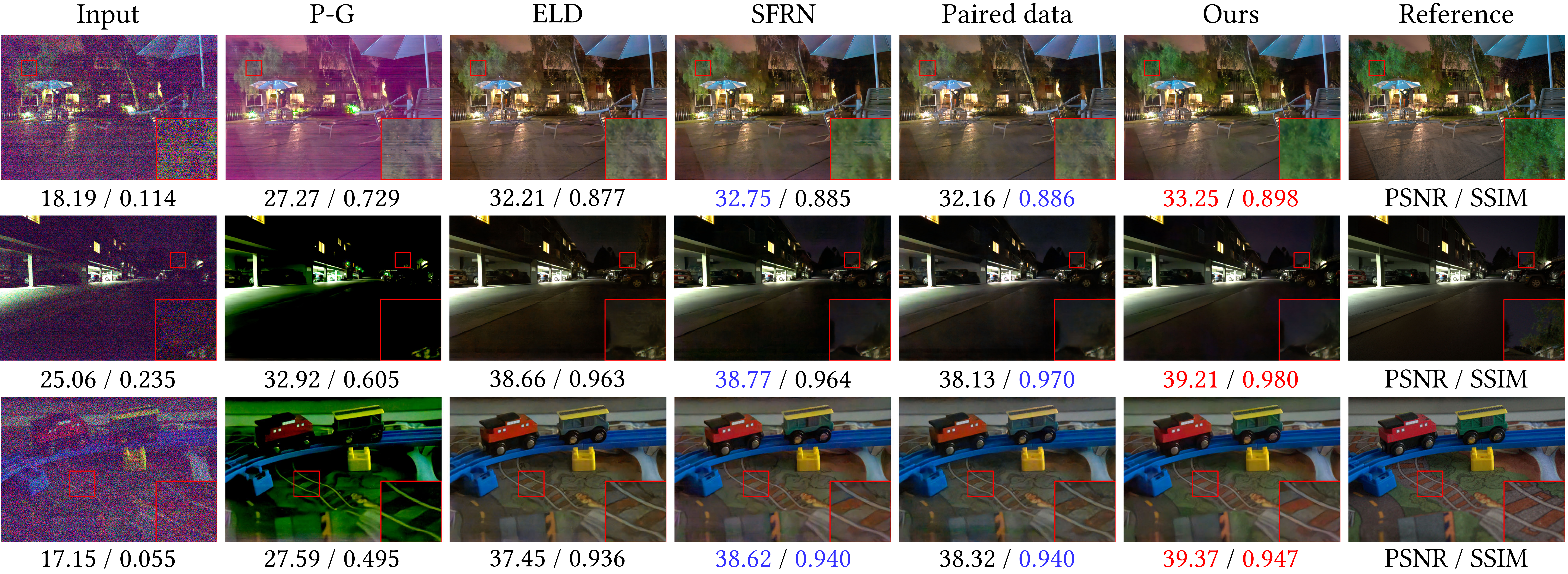}
        \caption{Raw image denoising results on images from the SID dataset. (Best viewed with zoom)}
        \label{fig:SID-compare}
    \end{figure*}
    

  \subsection{Results on ELD and SID Datasets}
    Table~\ref{tab:compare} summarizes the denoising performances over different exposure ratios of the denoising model trained with the data coming from different data formation models. Fig.~\ref{fig:ELD-compare} and Fig.~\ref{fig:SID-compare} have shown the comparisons on ELD dataset and SID dataset respectively.

    The model trained with synthetic data cannot completely remove the complicated real noise. The P-G model is far from the real noise distribution, so the performance is underdeveloped. ELD considered FPN and BLE, but it still deviates from the noise model, so there is color bias and residual noise. Although SFRN sampled real read noise, the patch-wise method cannot address the problem of dark shading inherently. The paired real data, despite containing real noise, is so fragile in learnability that the model cannot learn the accurate data mapping at all. By applying our learnability enhancement strategy to paired real data, the denoising performance gets a significant improvement in both quantitative results and visual comparison. As shown in Fig.~\ref{fig:ELD-compare} and Fig.~\ref{fig:SID-compare}, our method can provide clean denoising results with the clearest texture and the most exact color.

    The backbones of these models are all consistent, and their training settings are similar, so their performance differences only depend on the training data.
    The final performance of the model is significantly impacted by changes in the learnability of the data mapping.

    The performance of synthetic data is gradually improving as the noise modeling methods approximate the real noise model.
    SFRN addressed the limited data volume from the standpoint of synthetic data. The increase in learnability brought about by the growth of the data volume led to its success.
    Our learnability enhancement strategy proposes a different approach, not only increasing the limited data volume through SNA but also reducing the noise complexity through DSC. Proposing a comprehensive solution for learnability enhancement on paired real data enables our method to surpass previous work.
    
    \begin{figure*}[t!]
        \includegraphics[width=\textwidth]{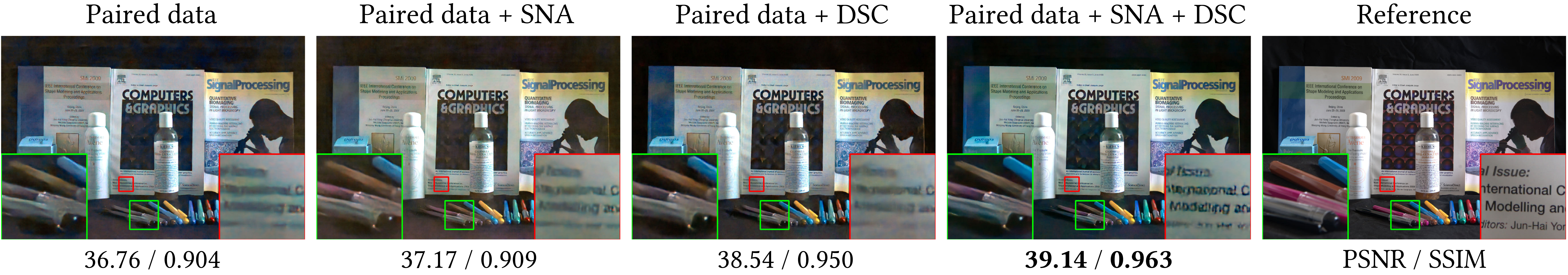}
        \caption{Visual results comparison of different training schemes. (Best viewed with zoom)}
        \label{fig:ablation}
    \end{figure*}
    \begin{figure*}[t!]
        \includegraphics[width=\textwidth]{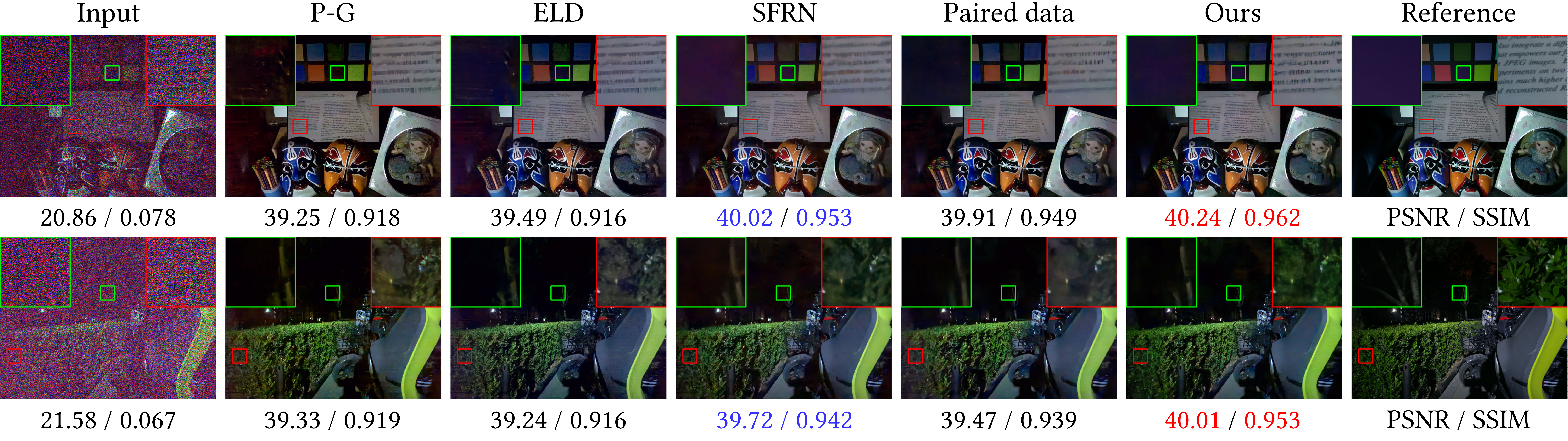}
        \caption{Raw image denoising results on low-light images captured by a RedMi K30 camera. (Best viewed with zoom)}
        \label{fig:IMX686}
    \end{figure*}

  \subsection{Ablation Study}
    \begin{table}[t!]
    \caption{Quantitative Results on Sony set of the ELD and SID. The best results have been shown in \textbf{bold}.}
    \label{tab:ablation}{%
        \begin{tabular}{c|c|cc|ccc}
        \hline
         &  & \multicolumn{2}{c|}{ELD} & \multicolumn{3}{c}{SID} \\ \cline{3-7}
        \multirow{-2}{*}{Method} & \multirow{-2}{*}{Index} & $\times 100$ & $\times 200$ & $\times 100$ & $\times 250$ & $\times 300$ \\ \hline
         & PSNR & {44.47} & {41.97} & {42.06} & {39.60} & {36.85} \\
        \multirow{-2}{*}{Paired data} & SSIM & {0.968} & {0.928} & {0.955} & {0.938} & {0.923} \\ \hline
         & PSNR & {44.91} & {42.68} & {42.36} & {39.63} & {37.02} \\
        \multirow{-2}{*}{\begin{tabular}[c]{@{}c@{}}Paired data\\ + SNA\end{tabular}} & SSIM & {0.976} & {0.941} & {0.955} & {0.937} & {0.920} \\ \hline

         & PSNR & {46.20} & {44.32} & {43.06} & {40.65} & {37.64} \\
        \multirow{-2}{*}{\begin{tabular}[c]{@{}c@{}}Paired data\\ + DSC\end{tabular}} & SSIM & {0.982} & {0.971} & {\textbf{0.960}} & {0.945} & {0.932} \\ \hline
         & PSNR & {\textbf{46.50}} & {\textbf{44.51}} & {\textbf{43.16}} & {\textbf{40.92}} & {\textbf{37.77}} \\
        \multirow{-2}{*}{\begin{tabular}[c]{@{}c@{}}Paired data\\+SNA+DSC\end{tabular}} & SSIM & {\textbf{0.985}} & {\textbf{0.973}} & {\textbf{0.960}} & {\textbf{0.947}} & {\textbf{0.934}} \\ \hline
        \end{tabular}%
        
        }
    \end{table}

    To study the contribution of each module we proposed, we compare the performance of models trained without SNA or DSC respectively.
    Table~\ref{tab:ablation} summarizes the quantitative results of the networks trained with the data from different training schemes. Fig.~\ref{fig:ablation} shows a representative comparison of our ablation study.

    Paired real data without any learnability enhancement appears blurry and noisy. The fragile learnability makes it difficult to learn an accurate mapping from a noisy image to its clean counterpart.
    
    SNA brings a small improvement in quantitative results while it promotes the denoised images with clear texture.
    The improvement of visual quality is attributed to the more precisely fitting brought about by the growth of data volume.
    However, the mapping dilemma still exists, which induces the model to overfit a biased mapping under the augmentation of SNA.
    
    DSC brings a large improvement in quantitative results while it promotes the denoised images with exact color.
    The improvement of quantitative results is attributed to the reduction of mapping complexity brought about by the correction of dark shading.
    However, DSC cannot significantly improve the denoising precision, which needs to rely on more precise fitting.
    
    The best performance is achieved using our learnability enhancement strategy. SNA focuses on improving the mapping precision while DSC focuses on reducing the mapping complexity. The combination of DSC and SNA constitutes our learnability enhancement, and it achieves the best quantitative results and visual quality.

  \subsection{Results on Real Imaging Scenarios}
    Since our method depends on the paired real data, we collected a small low-light denoising dataset referring the existing works~\cite{CVPR17/DND,CVPR18/SID,CVPR18/SIDD,TPAMI21/ELD,ECCV20/Yuzhi} by RedMi K30, a smartphone with an IMX686 sensor, to evaluate the generalization of our method.
    All noisy images were collected at maximum analog gain corresponding to ISO-6400. 
    The images shown in Fig.~\ref{fig:IMX686} are equivalent to the digital gain being set to 8.
    Fig.~\ref{fig:IMX686} has shown the comparisons on real imaging scenarios.
    The results are consistent with our comparative experiments on the SID dataset and ELD dataset. Our method still achieve the best denoising results with the clearest texture and the most exact colors, which demonstrate the high generalization of our method.

\section{Conclusion}
    In this paper, we introduce a new perspective to reform paired real data according to noise modeling.
    Our learnability enhancement inherently addresses the learnability bottleneck of paired real data through noise model decoupling.
    Two efficient techniques: shot noise augmentation and dark shading correction help the network efficiently learn the data mapping by increasing the data volume and reducing the noise complexity respectively.
    Extensive experiments on the public datasets and real imaging scenarios collectively demonstrates the state-of-the-art performance of our method.



\begin{acks}
This work is supported by the \grantsponsor{GS501100001809}{National Natural
Science Foundation of
China}{https://doi.org/10.13039/501100001809} (\grantnum{GS501100001809}{62131003},~\grantnum{GS501100001809}{62072038}). Part of this work is done during the internship of Hansen Feng at Megvii Technology.
\end{acks}

\bibliographystyle{ACM-Reference-Format}
\bibliography{Reference}

\end{document}